\documentclass[runningheads]{llncs}
\usepackage{graphicx}
\usepackage{cite}

\begin{document}
\title{Concept Extraction Using Pointer--Generator Networks}
%
%
\author{Alexander Shvets\inst{1}\orcidID{0000-0002-8370-2109} \and
Leo Wanner\inst{1,2}\orcidID{0000-0002-9446-3748}}
\authorrunning{A. Shvets and L. Wanner}
%
\institute{NLP Group, Pompeu Fabra University, Roc Boronat, 138, Barcelona, Spain \\
\email{\{alexander.shvets,leo.wanner\}@upf.edu} \and
Catalan Institute for Research and Advanced Studies (ICREA)}
\maketitle              
\begin{abstract}
Concept extraction is crucial for a number of downstream applications. However, surprisingly enough, straightforward single token/nominal chunk--concept alignment or dictionary lookup techniques such as DBpedia Spotlight still prevail. We propose a generic open-domain OOV-oriented extractive model that is based on distant supervision of a pointer--generator network leveraging bidirectional LSTMs and a copy mechanism. The model has been trained on a large annotated corpus compiled specifically for this task from 250K Wikipedia pages, and tested on regular pages, where the pointers to other pages are considered as ground truth concepts. The outcome of the experiments shows that our model significantly outperforms standard techniques and, when used on top of DBpedia Spotlight, further improves its performance. The experiments furthermore show that the model can be readily ported to other datasets on which it equally achieves a state-of-the-art performance.

\end{abstract}

\section{Introduction}
\label{sec:intro}

In knowledge discovery and representation, the notion of {\it concept} is most often used to refer to {\it sense}, i.e., `abstract entity' or `abstract object' in the Fregean dichotomy of {\it sense} vs. {\it reference} \cite{Frege1892}. In Natural Language Processing (NLP), the task of {\it Concept Extraction} (CE) deals with the identification of the language side of the concept coin, i.e., Frege's {\it reference}. Halliday \cite{Halliday13} offers a syntactic interpretation of {\it reference}. In his terminology, it is a ``classifying nominal group''. For instance, {\it renewable energy} or {\it nuclear energy} are classifying nominal groups: they denote a class (or type) of energy, while, e.g., {\it cheap energy} or {\it affordable energy} are not: they do not typify, but rather qualify {\it energy} (and are thus ``qualifying nominal groups'').

CE is crucial for a number of downstream applications, including, e.g., language understanding, ontology population, semantic search, and question answering; it is also the key to entity linking \cite{logeswaran-etal-2019-zero}. In generic open domain subject-neutral discourse across different (potentially unrelated) subjects, indexing the longest possible nominal chunks and their head words located in sequences of tokens between specified ``break words" \cite{woods1997conc} and special dictionary lookups such as {\it DBpedia Spotlight} \cite{isem2013daiber} and {\it WAT} \cite{Piccinno:2014:TWN:2633211.2634350} are very common techniques. They generally reach outstanding precision, but low recall due to constant evolvement of the language vocabulary. Advanced deep learning models that already dominate CE in specialized closed domain discourse on one or a limited range of related subjects, e.g., biomedical discourse
\cite{habibi2017deep,Tulkens-etal19}, and that are also standard in keyphrase extraction \cite{meng2017deep,Al-Zaidy-etal19} are an alternative. However, such models need a tremendous amount of labeled data for training.

We present an operational CE model that utilizes pointer--generator networks \cite{see2017get} and bidirectional long short-term memory (LSTM) units \cite{graves2005framewise} to retrieve concepts from general discourse textual material.\footnote{We adopt Halliday's notion of classifying nominal group as definition of a concept.}
Furthermore, since for a generic, domain-independent concept extraction model we need a sufficiently large training corpus that covers a vast variety of topics and no such annotated corpora are available, we opt for distant supervision to create a sufficiently large and diverse dataset. Distant supervision consists in automatic labeling of potentially useful data by an easy-to-handle (not necessarily accurate) algorithm to obtain an annotation which is likely to be noisy but, at the same time, to contain enough information to train a robust model \cite{mintz2009distant}. Two labeling schemes are considered. Experiments carried out on a dataset of 250K+ Wikipedia pages show that copies of our model trained differently and joined in an ensemble significantly outperform standard techniques and, when used on top of DBpedia Spotlight, further improve its performance by nearly 10\%.

\section{Related work}
\label{sec:relwork}

In this section, we focus on the review of generic discourse CE; for a comprehensive review of the large body of work on specialized discourse CE, and, in particular, on biomedical CE;
see, e.g., \cite{Hailu19}. We also do not discuss recent advances in keyphrase extraction \cite{Al-Zaidy-etal19} because their applicability to generic concept extraction is limited due to specificity of the task.

The traditional CE techniques interpret any single and multiple token nominal chunk as a concept \cite{woods1997conc} or do a dictionary lookup, as, e.g., {\it DBpedia Spotlight} \cite{isem2013daiber}, which matches and links identified nominal chunks with DBpedia entries (6.6M entities, 13 billion RDF triples)\footnote{https://wiki.dbpedia.org/develop/datasets/dbpedia-version-2016-10}, based on the Apache OpenNLP\footnote{https://opennlp.apache.org/} models for phrase chunking and named entity recognition (NER).
Given the large coverage of DBpedia, the performance of DBpedia Spotlight is rather competitive.
However, obviously, the presence of an entry cannot always be ensured. Consider, e.g., a paper title ``Natural language understanding with Bloom embeddings, convolutional neural networks and incremental parsing'', where DBpedia Spotlight does not detect ``Bloom embeddings'' or ``incremental parsing'', as there are no such entries in DBpedia.

As DBpedia Spotlight, AIDA \cite{yosef2011aida} relies on an RDF repository, YAGO2. WAT and its predecessor TagMe \cite{Piccinno:2014:TWN:2633211.2634350} use a repository of possible spots made of wiki-anchors, titles, and redirect pages. Both TagMe and WAT rely on statistical attributes called {\it link probability} and {\it commonness}; WAT draws furthermore on a set of statistics to prune a set of mentions using an SVM classifier. Wikifier \cite{cheng2013relational} focuses on relation extraction, relying on a NER, which uses gazetteers extracted from Wikipedia and simple regular expressions to combine several mentions into a single one. All of them are used for state-of-the-art entity linking and (potentially nested) entity mention detection and typing \cite{zhang-etal-2019-ernie,hasibi2015entity}.
FRED \cite{gangemi2017semantic} also focuses on extraction of relations between entities, with frames \cite{FillmoreBaker01} as the underlying theoretical constructs. Unlike Wikifier and FRED, e.g., OLLIE \cite{mausam-etal-2012-open} does not rely on any precompiled repository. It outperforms its strong predecessors REVERB \cite{fader-etal-2011-identifying} in relation extraction by expanding the set of possible relations and including contextual information from the sentence from which the relations are extracted.

A number of works focus on the recognition of named entities, which are the most prominent concept type. NERs work at a sentence level and aim at labeling all occurred instances. Among them, Lample et al. \cite{lample2016neural} provide a state-of-the-art NER model that avoids traditional heavy use of hand-crafted features and domain-specific knowledge. The model is based on bidirectional LSTMs and Conditional Random Fields (CRFs) that rely on two sources of information on words: character-based word representations learned from an annotated corpus and unsupervised word representations learned from unannotated corpora.
Another promising approach to NER is fine-tuning of a language representation model such as, e.g., BERT \cite{devlin-etal-2019-bert}. The pre-trained BERT model can be fine-tuned with just one additional output layer to create state-of-the-art models for a wide range of tasks, including NER, without substantial task-specific architecture modifications.

\section{Description of the model}
\label{sec:desc_model}

We implement a deep learning model and a large-scale annotation scheme for the distant supervision to cope autonomously with dictionary-independent generic CE and to a possible extent complement present lookup-based approaches to increase their recall. In addition, we would like our model to perform decently on pure NER tasks with a small gap to models specifically tuned for the NER datasets. The model follows the well-established tendency in information extraction adopted for NER and extractive summarization and envisage CE as an attention-based sequence-to-sequence learning problem.

\begin{figure*}[t]
	\centering
	\includegraphics[width=1.0\textwidth]{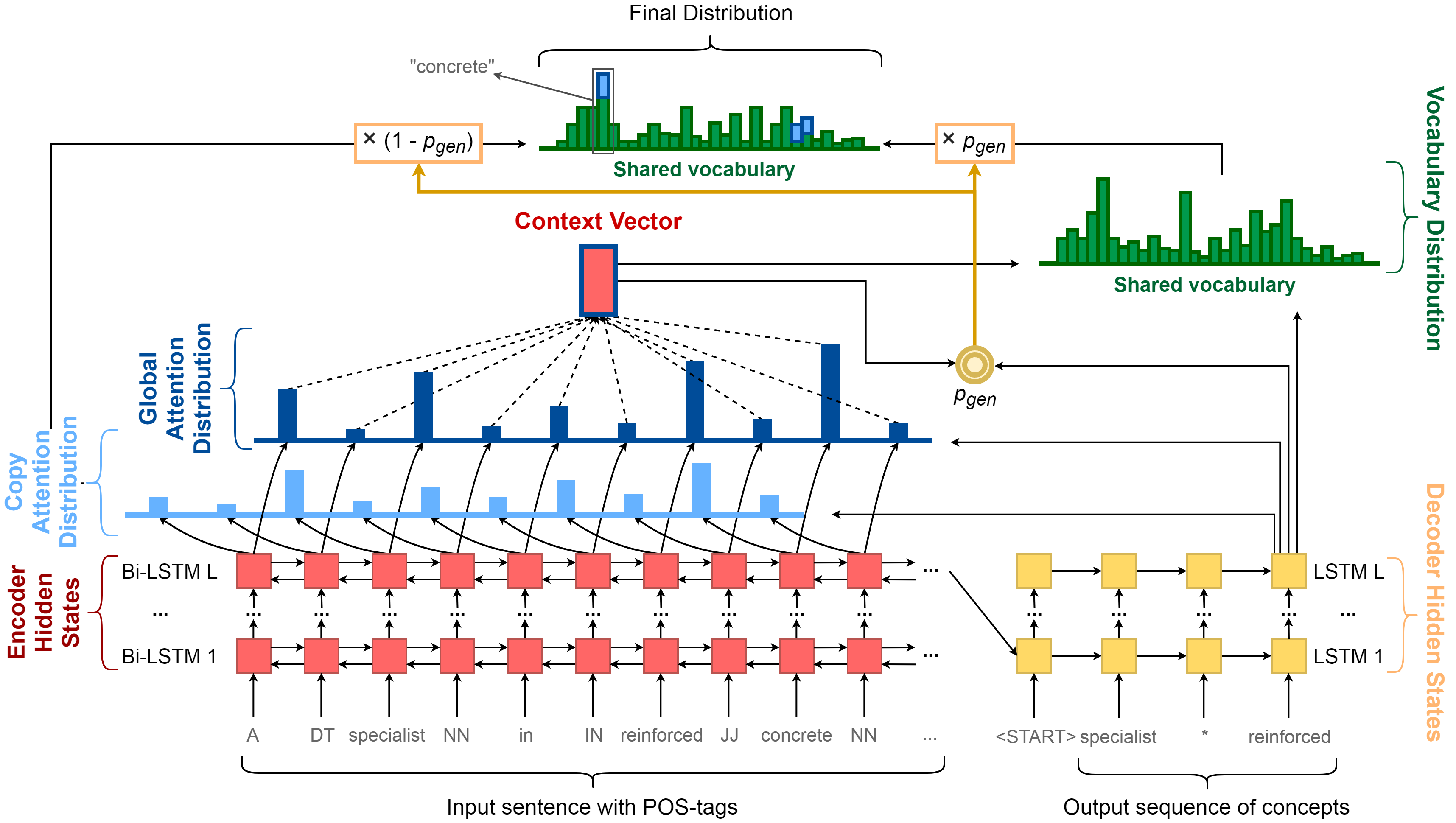}
	\caption{The neural architecture for concept extraction}
	\label{fig:model}
\end{figure*}

\subsection{Overview of the model}
\label{subsec:model-overview}

As a basis of our model, we use the pointer--generator network proposed in \cite{see2017get} that aids creation of summaries with accurate reproduction of information. In each generation step $t$, the \textit{pointer} allows for copying words $w_{i}$ from the source sequence to the target sequence using distribution of attention layer $a^{t}$, while the \textit{generator} samples tokens from the learned vocabulary distribution ${P}_{vocab}$, conditioned by a context vector $h_{t}^{*}$ produced by the same attention layer which is built based on hidden states $h_{i}$ of an encoder and states $s_{t}$ of a decoder (in each case, a bidirectional LSTM \cite{graves2005framewise}). In addition, coverage mechanism is applied to modify $a^{t}$ using a coverage vector $c^{t}$ to avoid undesirable repetitions in the output sequence. Specifically, to produce a word $w$, the above-mentioned distributions are combined into a single final probability distribution being weighted using the \textit{generation probability} ${p}_{gen}$$\in$[0,1]:\begin{equation}
P(w) = {p}_{gen}{P}_{vocab}(w) + (1 - {p}_{gen}){\sum}_{i:w_{i}=w}{a_{i}^{t}}
\end{equation}
where ${P}_{vocab}(w)$ is the vocabulary distribution, which is zero if $w$ is an out-of-vocabulary (OOV) word; $a^{t}$ is the attention distribution; $w_{i}$ - tokens of the input sequence; ${\sum}_{i:w_{i}=w}a_{i}^{t}$ is zero if $w$ does not appear in the source sequence. According to \cite{see2017get}, individual vectors, distributions, and probability ${p}_{gen}$ are defined as follows:
\begin{equation}
c^{t} = {\sum}_{t'=0}^{t-1}a^{t'}
\end{equation}
\begin{equation}
e_{i}^{t} = v^{T}tanh(W_{h}h_{i}+W_{s}s_{t}+w_{c}c_{i}^{t}+b_{attn})
\end{equation}
\begin{equation}
a^{t} = softmax(e^{t})
\end{equation}
\begin{equation}
h_{t}^{*} = {\sum}_{i} {a_{i}^{t}h_{i}}
\end{equation}
\begin{equation}
P_{vocab} = softmax(V'(V[s_{t},h_{t}^{*}]+b)+b')
\end{equation}
\begin{equation}
p_{gen} = {\sigma}(w_{h*}^{T}h_{t}^{*}+w_{s}^{T}s_{t}+w_{x}^{T}x_{t}+b_{ptr})
\end{equation}
where $v$, $W_{h}$, $W_{s}$, $w_{c}$, $b_{attn}$, $V$, $V'$, $b$, $b'$, $w_{h*}$, $w_{s}$, $w_{x}$, $b_{ptr}$ are learnable parameters, $T$ stands for the transpose of a vector, $x_{t}$ is the decoder input, and ${\sigma}$ is the sigmoid function.

To adapt this basic model to the task of CE, we applied several modifications to it (cf., Figure \ref{fig:model}\footnote{We use a similar layout as in \cite{see2017get} for easier comparison of our extension with the original model.}): (i) following Gu et al. \cite{gu2016incorporating}, we use separate distributions for copy attention and general attention, instead of one for both; (ii) experiments have shown that encoders and decoders with several LSTM layers perform better than with a single layer, such that we work with multiple layer LSTMs; how many is determined using a development dataset; (iii) we adapt the forms of input and target sequences to the specifics of the task of CE. The input is comprised of tokens and their part-of-speech (PoS) tags (e.g., `The DT President NN is VBZ elected VBD by IN a DT direct JJ vote NN'). The target sequence concatenates concepts in the order they appear in the text and separates them by a token ``*'' especially introduced to partition the output (e.g., `President * direct vote').

This model is naturally applicable to the task of CE since it facilitates the selection and transfer of subsequences of tokens (= concepts) from a given source sequence of tokens (= text input) to the target sequence (= partitioned sequence of concepts). The pointer mechanism implies the ability to cope with OOV words, which is crucial for universal CE, while the generator implies the ability to adjust vocabulary distribution for selecting the next word (which might be a termination token ``*'') based on a given context vector, which allows us to implicitly take into account the domain specifics and linguistic features that facilitate the task of CE. Furthermore, the updating of vocabulary distribution adds the possibility to vanish or strengthen the copy effect and thus learn to distinguish concepts with outer modifiers (such as, e.g.,``hot \textit{air}'', ``[fully] crewed \textit{aircraft}'', ``reinforced \textit{group}'') from multiword concepts (such as, e.g., ``\textit{hot air balloon}'', ``\textit{unmanned aerial vehicle}'', ``\textit{reinforced concrete}'').

\subsection{Training and applying the model}
\label{subsec:tr-ap}
For training, token sequences are taken from annotated sentences (see the compilation of the annotated training dataset in Section \ref{subsec:compil} below) with a sliding overlapping window of a fixed maximum length (see the Experiments section), which is minimally expanded if needed in order not to deal with incomplete concepts at the borders. The trained model is applied to unseen sentences, which are also split into sequences of tokens with an overlapping window of the same size, without any expansion. Finally, the corresponding mentions in the plain text are determined since the output format does not include offsets. In particular, following \cite{hasibi2015entity}, we find all possible matches for all detected concepts and then successively select non-nested concepts from the beginning to the end of the sentence, giving priority to the longest, in case of a multiple choice.

\section{Datasets}
\label{sec:datasets}

In what follows, we describe the data and the procedure for their weak annotation to create extensive training and test datasets.

\subsection{Data}
\label{subsec:data}
\begin{figure}[ht]
	\centering
	\hrule
	\includegraphics[width=0.7\columnwidth]{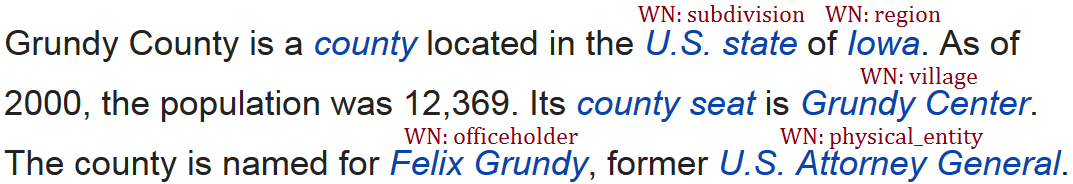}
	\hrule
	\caption{Ground truth concept annotation}
	\label{fig:gtruth}
\end{figure}

We take a snapshot of the WordNet synset-typed\footnote{https://wordnet.princeton.edu/} Wikipedia \cite{schenkel2007yawn}, from which we use the raw texts of the Wikipedia pages and text snippets of the links to other pages as ground truth concepts regardless their type;
cf., Figure \ref{fig:gtruth}\footnote{Wikipedia does not contain self-links, therefore the concept {\it ``Grundy County''} in a text from the self-titled page is not a link.}. These links often share the headings of anchor pages, which are in most cases some real-world entities, cf., e.g., ``Arthur Heurtley House'', ``Price Tower'', etc. Sometimes, they are also lexical variations of terms behind the link, as, e.g., the highlighted link in the fragment ``the two small {\it coastal battleships} General-Admiral Graf Apraxin and Admiral Senyavin'' leads to the page named ``Coastal defence ship''.

\noindent The manual annotation of multi-word expressions in 100 randomly selected sentences with at least one multi-word link in each by a professional linguist showed that at least 63\% of such phrases are indeed concepts (cf., e.g., ``punctuated equilibrium'', ``chief of staff'', ``2004 presidential election''). For our work, we selected several data subsets from the collection of Wikipedia pages: 250 K pages to be weakly, but {\it densely} annotated.\footnote{Henceforth, we refer to the link snippet-based annotation of the pages as a {\it sparse} gold standard annotation since it covers by far not all concepts encountered in a page. Our distant supervision-based annotation is referred to as {\it dense} annotation since it (supposedly) covers all concepts on a given page. As usual, distant supervision-based annotation is also referred to as {\it weak} since it is an automatic annotation.} Out of these 250K pages, 220K are used for training and 30K for validation.
In addition, we use 7K Wikipedia pages with the sparse gold standard annotation as development set for choosing parameters of distant supervision and selecting the best model among several models trained with different parameters, and 7K pages with the sparse gold standard annotation as test set. 

\subsection{Compilation of the training corpus}
\label{subsec:compil}

\begin{figure}[t]
	\centering
	\includegraphics[width=1.0\columnwidth]{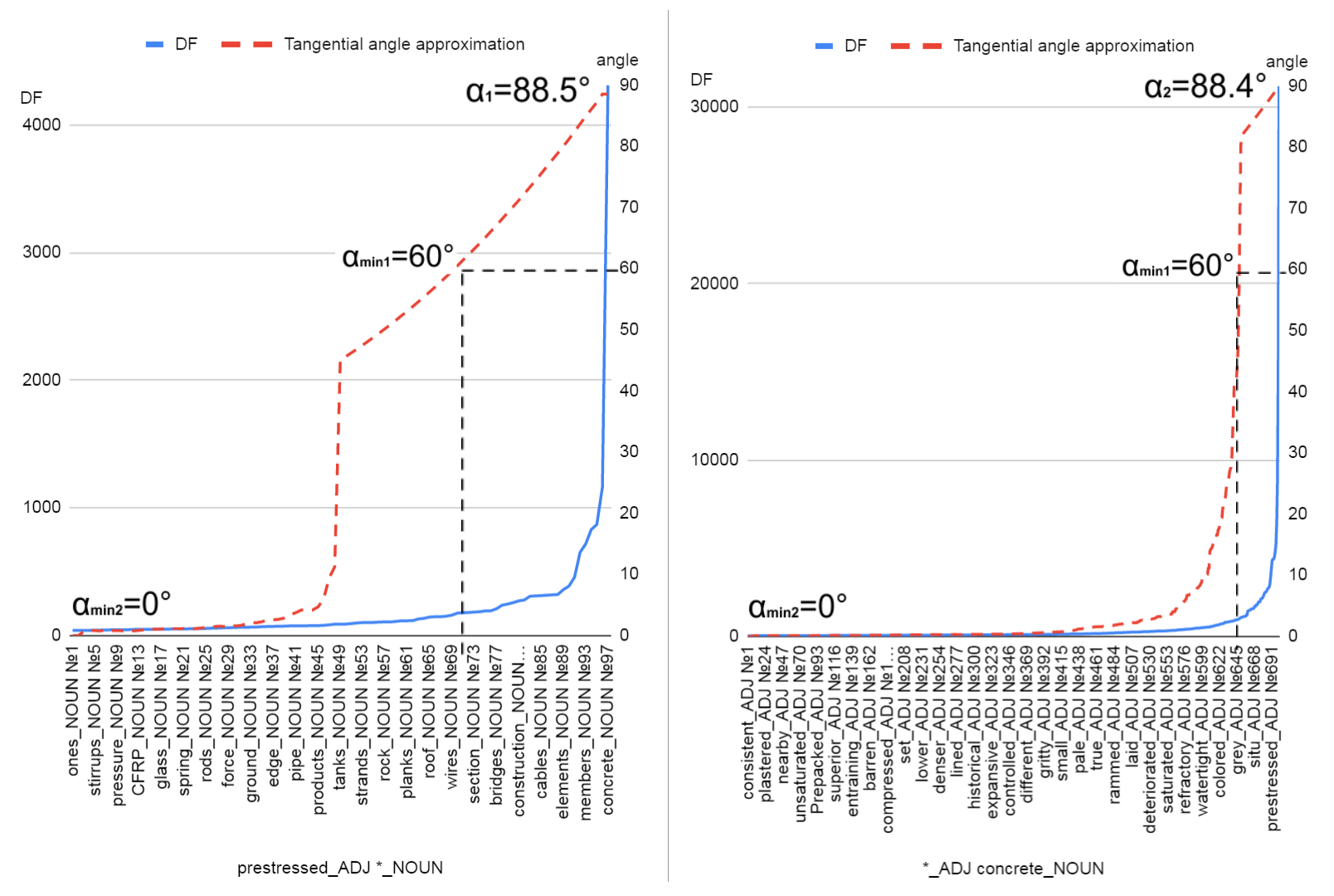}
	\caption{Relation between document frequency and coarse-grained tangential angle approximation}
	\label{fig:angles}
\end{figure}

We automatically create a (noisy) training corpus using two various annotators over a large unlabeled dataset: DBpedia Spotlight with the value of its confidence coefficient that gains the highest recall and our own algorithm that uses a number of rules and heuristics. 
Our labeling is based on the sentence-wise analysis of statistical and linguistic features of sequences of tokens. First, named entities and multiple token concepts and then single token concepts are identified. The algorithm covers the following tasks:

\noindent {\bf 1. Application of a statistical NER model.} A significant number of concepts in Wikipedia are capitalized terms, which can be captured by statistical named entity recognizers (NER); see the Related Work section above. Therefore, at first, SpaCy's state-of-the-art NER model \cite{honnibal2017spacy} is applied with a successive elimination of used tokens for further processing. The next steps are applied then separately to fragments of texts located between the identified NEs.\\
\noindent {\bf 2. Selection of $n$-grams as fragments of NP chunks that can form part of multiple token concepts.} For this task, we formed the PoS-patterns based on Penn Treebank tagset\footnote{https://www.ling.upenn.edu/courses/Fall\_2003/ling001/\linebreak penn\_treebank\_pos.html}, which were inherited from the patterns for multiword expression detection introduced in \cite{cordeiro2016ufrgs} and expanded here resulting in the following set:
$P$ = \{N\_N, J\_N, V\_N, N\_J, J\_J, V\_J, N\_of\_N, N\_of\_DT\_N, N\_of\_J, N\_of\_DT\_J, N\_of\_V, N\_of\_DT\_V, CD\_N, CD\_J\},
\noindent where N stands for ``noun'', i.e., NN$\mid$NNS$\mid$NNP$\mid$NNPS, J stands for ``adjective'', i.e., JJ$\mid$JJR$\mid$JJS, V - ``verb'' but limited to VBD$\mid$VBG$\mid$VN, CD - ``cardinal number'', DT - ``determiner'', and ``of'' is an exact pronoun. Each pattern matches an $n$-gram with two open-class lexical items and at most two auxiliary tokens between them.\\
{\bf 3. Assessment of the distinctiveness of each selected $n$-gram.} The distinctiveness of the selected $n$-grams is assessed using word co-occurrences from the Google Books Ngram Corpus \cite{lin-etal-2012-syntactic}. Let us assume a given $n$-gram $T_1A_1A_2T_2 \in c_k$, where $T_1$ and $T_2$ are open class lexical items and $A_1$ and $A_2$ are optional auxiliary tokens, and $c_k$ is a set of all $n$-grams of a particular kind of pattern $p_k \in P$. We use $T_1A_1A_2T_2$ as a point of a function that passes through normalized document frequencies of a set of similar $n$-grams $T_1A_1A_2T_j$, $j$ $\in$ \{$i$ $|$ $T_1A_1A_2T_i \in c_k$\} arrayed in ascending order, to find a tangential angle at this point $\alpha_1 \in [0^{\circ}; 90^{\circ})$. Similarly, $\alpha_2 \in [0^{\circ}; 90^{\circ})$, is a tangential angle at the point $T_1A_1A_2T_2$ on a curve of ordered frequencies of $n$-grams $T_hA_1A_2T_2$, $h$ $\in$ \{$i$ $|$ $T_iA_1A_2T_2 \in c_k$\}.
We leverage these angles to check how prominent an $n$-gram is, i.e., to what extent it differs from its neighbors by overall usage. In case an $n$-gram is located among equally prominent $n$-grams with a tangential angle close to $0^{\circ}$, we do not consider it as a potential part of a concept since it does not show a notable distinctiveness inherent in concepts, especially in common idiosyncratic concepts. The thresholds $\alpha_{min_1}$ and $\alpha_{min_2}$ ($\alpha_{min_1}$ $\geq$ $\alpha_{min_2}$) for minimally allowed tangential angles such as $\max(\alpha_1, \alpha_2)$ $\geq$ $\alpha_{min_1}$, $\min(\alpha_1, \alpha_2)$ $\geq$ $\alpha_{min_2}$ are predermined in development experiments.
We calculate tangential angles through central difference approximation with a coarse-grained grid:
\begin{equation}
\alpha = \arctan(\frac{f(x+h) - f(x-h)}{2h}) \cdot \frac{180}{\pi}
\end{equation}
\noindent where $h$ was chosen large enough ($h=50$ in general, and it is maximum possible on the borders) for smoothing the curve to eliminate numerous abrupt changes in document frequency with relatively low amplitude. Thus, the approximation is intentionally carried out less accurately to result in such values that in practice form a curve with longer monotonous sections; (cf., Figure \ref{fig:angles}) for an example of assessing the prominence of an $n$-gram ``prestressed concrete'', i.e., in the above notation, $T_1$ equals ``prestressed\_ADJ'', $A_1$ and $A_2$ are omitted, and $T_2$ equals ``concrete\_NOUN''.

\begin{table}
	\centering
	{\small
	\caption{\label{tab:candidate_angles}Tangential angles of concept candidates}

	\begin{tabular}{|l|c|c|}
		\hline
		Candidate & Angle 	& Wiki-term	\\\hline
		reinforced\_ADJ concrete\_NOUN &	89.77	& YES \\
		mixed\_ADJ concrete\_NOUN	&	89.07	& NO \\
		prestressed\_ADJ concrete\_NOUN &	88.40	& YES \\
		pre-cast\_ADJ concrete\_NOUN &	83.66	& YES \\
		first\_ADJ concrete\_NOUN &	33.12	& NO \\
		original\_ADJ concrete\_NOUN &	16.63	& NO \\
		massive\_ADJ concrete\_NOUN &	9.85	& NO \\
		resistant\_ADJ concrete\_NOUN &	8.08	& NO \\
		special\_ADJ concrete\_NOUN &	5.66	& NO \\
		polymer\_ADJ concrete\_NOUN &	4.03	& YES \\
		tall-wall\_ADJ concrete\_NOUN	& 1.90	& NO \\
		large\_ADJ concrete\_NOUN	&	1.75	& NO \\
		open\_ADJ concrete\_NOUN	&	0.75	& NO \\
		\ldots	&	\ldots	& \ldots \\
		unusual\_ADJ concrete\_NOUN	& OOV	& NO \\
		raised\_ADJ concrete\_NOUN		& OOV	& NO \\
		\ldots	& \ldots	& \ldots \\
		\hline
	\end{tabular}
	}
\end{table}

Table \ref{tab:candidate_angles} illustrates how the approximations of tangential angles differentiate classifying nominal groups from qualifying nominal groups. The most of the candidates with a large tangential angle have a separate article in Wikipedia (i.e., they are likely to be concepts) while candidates with a small tangential angle or without an entry in Google Books (OOV) do not have a Wikipedia article in general. This shows that the chosen criterion for differentiating the concepts is suitable for weak annotation within distant supervision.


Grid search was applied to find the best combination of parameters $\alpha_{min_1}$ and $\alpha_{min_2}$ from the three possible tangential angles corresponding to the different levels of the distinctiveness of a concept: 85$^{\circ}$, 60$^{\circ}$, and 0$^{\circ}$. As a result, $\alpha_{min_1} = 60^{\circ}$ and $\alpha_{min_2} = 0^{\circ}$ gave the best scores on the development set and were used for annotation of the training set.\\
\noindent {\bf 4. Combination of intersected highly distinctive parts as concepts.} We combine those distinctive $n$-grams that share common tokens and iteratively drop the last token in each group if it is not a noun, in order to end up with complete NP candidate concepts (e.g., ``value of the played card'' is a potential concept corresponding to the patterns \{N\_of\_DT\_V; V\_N\}). Some single-word concepts already might appear at this point.\\
{\bf 5. Recovery of missed single-word concepts.}
To enrich the set of candidate concepts, we consider all unused nouns and numbers in a text as single-word concept candidates.\\
The obtained training corpus contains moderate amount of noise: the proposed annotation algorithm outperformes some baselines and might be used for CE by itself (cf. setup (A) in Tables \ref{tab:eval_small} and \ref{tab:eval_large} with results of evaluation in the following section).

\section{Experiments}
\label{sec:exp}

\subsection{Setup of the experiments}
\label{subsec:setup}

For our experiments, we use the realization of See et al. \cite{see2017get}'s pointer--generator model in the OpenNMT toolkit \cite{klein2018opennmt}, 
which allows for the adaptation of the model to the task of CE along the lines described in Section \ref{subsec:model-overview} above. Instead of the attention mechanism used in \cite{see2017get}, we use the default OpenNMT attention \cite{luong2015effective} since it showed to perform better.
The model has 512-dimensional hidden states and 256-dimensional word embeddings shared between encoder and decoder. We use a vocabulary of 50k words as we rely mostly on a copying mechanism which uses dynamic vocabulary made up of words from the current source sequence. We train using the Stochastic Gradient Descent on a single GeForce GTX 1080 Ti GPU with a batch size of 64.
We trained the CE-adapted pointer--generator networks of two and three bi-LSTM layers with 20K and 100K training steps on the two training datasets (obtained using Google Books and DBpedia Spotlight, respectively; see above). Validation and saving of checkpoint models was performed at each one-tenth of the number of training steps.

In order to compare our extended pointer--generator model with state-of-the-art techniques, several efficient entity extraction algorithms were chosen as baselines: OLLIE \cite{mausam-etal-2012-open}, AIDA \cite{yosef2011aida}, AutoPhrase+ \cite{shang2018automated}, DBpedia Spotlight \cite{isem2013daiber}, WAT \cite{Piccinno:2014:TWN:2633211.2634350},\footnote{FRED \cite{gangemi2017semantic} was not used as baseline as it is not scalable enough for the task: its REST service has a strong limitation on a number of possible requests per day, and it fails on processing long sentences (approximately more than 40 tokens).} and several state-of-the-art NER models, namely SpaCy NER \cite{honnibal2017spacy}, FLAIR NER \cite{akbik-etal-2019-flair} and two deep learning-based NER models \cite{lample2016neural,devlin-etal-2019-bert}\footnote{https://github.com/glample/tagger}\footnote{https://github.com/kyzhouhzau/BERT-NER}.
AutoPhrase+ was used in combination with the StanfordCoreNLP PoS-tagger (as it was reported to show better performance with PoS-tags) and trained separately on its default DBLP dataset and on the above-mentioned raw Wikipedia texts our training dataset is composed of. Its output was slightly modified by removing auxiliary tokens from the beginning and the end of the phrase to make it more competitive with the rest of the algorithms. OLLIE's and SpaCy's outcomes were also modified the same way, which improved their performance. DBpedia Spotlight was applied with two different values of confidence coefficient: 0.5 (default value) and 0.1, which increases the recall.

The performance is measured in terms of precision, recall, and $F_{1}$-score, aiming at high recall, first of all. Since positive ground truth examples are sparse, and there are no negative examples, we treated only the detected concepts that partially overlapped the ground truth concepts as false positives. Concepts that have the same spans as the ground truth concepts are counted as true positives, and missed ground truth concepts as false negatives. This perfectly meets our goal to detect the exact match. It also allows us to penalize brute force high-recall algorithms that produce a large number of nested concepts, which are of limited use in real-world applications.

Table \ref{tab:eval_small} shows the reached performance on the domain-specific datasets, and Table \ref{tab:eval_large} on the open domain set. The sign ``*'' stands for modifications made on cutting some first and last words of detected concepts in order to present them as ``canonic'' noun phrases, and ``**'' stands for removing nested concepts when this procedure gave better scores. 

\begin{table}
	\centering
	\caption{\label{tab:eval_small}Results on the domain-specific datasets}
	\small
	\resizebox{\columnwidth}{!}{\begin{tabular}{|l|l|c|c|c|c|c|c|}
			\hline
			\multicolumn{2}{|c|}{} & \multicolumn{3}{c|}{``Architecture''} & \multicolumn{3}{c|}{``Terrorist groups''}\\\hline
			Setup & Model 	& $P$	& $R$	& $F_{1}$ & $P$ & $R$	& $F_{1}$\\\hline
			& FLAIR (Akbik et al., 2019) & 0.79 & 0.74 & 0.76 & 0.77 & 0.66 & 0.71\\
			& BERT NER (Delvin et al., 2019) & 0.78 & 0.74 & 0.76 & 0.78 & 0.67 & 0.72\\
			& AutoPhrase+$_{DBLP}^{**}$ (Shang et al., 2018) & 0.38 & 0.44 & 0.41 & 0.31 & 0.34 & 0.33\\
			& AutoPhrase+$_{WIKI}^{**}$ (Shang et al., 2018) & 0.42 & 0.52 & 0.46 & 0.36 & 0.45 & 0.40\\
			& SpaCy NER (Honnibal and Montani, 2017) & 0.59 & 0.51 & 0.55 & 0.5 & 0.41 & 0.45\\
			& SpaCy NER$^{*}$ (Honnibal and Montani, 2017) & 0.71 & 0.61 & 0.66 & 0.59 & 0.49 & 0.54\\
			& NER Tagger (Lample et al., 2016) & 0.78 & 0.71 & 0.75 & 0.76 & 0.65 & 0.7\\
			& WAT$^{**}$ (Piccinno and Ferragina, 2014) & 0.66 & 0.53 & 0.59 & 0.64 & 0.5 & 0.56\\
			& Spotlight$_{0.5}$ (Daiber et al., 2013) & {\bf 0.85} & 0.74 & 0.79 & {\bf 0.8} & 0.7 & \textbf{0.75}\\
			& Spotlight$_{0.1}$ (Daiber et al., 2013) & 0.7 & 0.79 & 0.74 & 0.65 & 0.77 & 0.7\\
			& OLLIE$^{*}$ (Schmitz et al., 2012) & 0.46 & 0.2 & 0.28 & 0.41 & 0.22 & 0.28\\		
			& AIDA (Yosef et al., 2011) & 0.76 & 0.57 & 0.65 & 0.74 & 0.54 & 0.62\\\hline
			(A) & DSA$_{(60,0)}$ & 0.63 & 0.74 & 0.68 & 0.5 & 0.64 & 0.56\\
			(B) & ${PG}_{(3L,80K)}(DSA_{DICT})$ & 0.67 & 0.77 & 0.72 & 0.61 & 0.73 & 0.66\\
			(C) & ${PG}_{(2L,18K)}(DSA_{(60,0)})$ & 0.7  & 0.8 & 0.75 & 0.59  & 0.72 & 0.65\\\hline
			(D) & (B) + (C) & 0.75 & 0.83 & 0.79 & 0.66 & 0.77 & 0.71\\
			(E) & (B) + (C) + Spotlight$_{0.1}$ & 0.78 & 0.85 & 0.81 & 0.7 & {\bf 0.8} & \textbf{0.75}\\
			(F) & (B) + (C) + Spotlight$_{0.5}$ & 0.78 & 0.85 & 0.81 & 0.7 & {\bf 0.8} & \textbf{0.75}\\
			(G) & (C)  + Spotlight$_{0.5}$ & 0.79 & {\bf 0.86} & \textbf{0.82} & 0.69 & 0.79 & 0.74\\
			\hline
	\end{tabular}}
\end{table}
	
Table \ref{tab:eval_large} displays the scores for two different experiment runs. In the first, only concepts with an assigned WordNet type label in our typed Wikipedia dataset (in their majority, named entities; cf. \cite{schenkel2007yawn} for details of the typification) were considered as positive examples (from about 276K nouns in the test set, only 83K nouns, i.e., about 30\%, were part of ground truth concepts); in the second, all text snippets of the links were taken as ground truth concepts (from about 390K nouns in the test set, 141K nouns, i.e., about 36\%, were part of ground truth concepts). Setups A -- H display the performance of different variants of our model. `$DSA$' stands for initial distant supervision annotation obtained using DBpedia Spotlight, i.e., a dictionary lookup, (`$DSA_{DICT}$'), and with the proposed token-cooccurrence frequency-based method (`$DSA_{(60,0)}$') (cf. Step 3 of the compilation of the training corpus), where the values in parentheses correspond to $\alpha_{min_1}$ and $\alpha_{min_2}$, which gave the best scores on the development set. ${PG}_{(2L,18K)}$ and ${PG}_{(3L,80K)}$ stand for pointer--generator networks with parameters shown in parentheses chosen using the development set (2 layers, $18K$/$20K$ training steps and 3 layers, $80K$/$100K$ training steps correspondingly).
	
\begin{table}
	\centering
	\caption{\label{tab:eval_large}Results on a large-scale open-domain dataset}
	\small
	\resizebox{\columnwidth}{!}{\begin{tabular}{|l|l|c|c|c|c|c|c|}
			\hline
			\multicolumn{2}{|c|}{} & \multicolumn{3}{p{2.5cm}|}{Only WordNet-typed concepts} & \multicolumn{3}{p{2.5cm}|}{All ground truth concepts}\\\hline
			Setup & Model 	& $P$	& $R$	& $F_{1}$	& $P$	& $R$	& $F_{1}$\\\hline
			& FLAIR (Akbik et al., 2019) & {\bf 0.8} & 0.74 & 0.77 & {\bf 0.79} & 0.59 & 0.67\\
			& AutoPhrase+$_{DBLP}^{**}$ (Shang et al., 2018)	& 0.42	& 0.45	& 0.43	& 0.4	& 0.43	& 0.41\\
			& AutoPhrase+$_{WIKI}^{**}$ (Shang et al., 2018)	& 0.46	& 0.52	& 0.49	& 0.43	& 0.49	& 0.46\\
			& NER Tagger (Lample et al., 2016) & 0.78 & 0.72 & 0.75 & 0.77 & 0.58 & 0.66\\	
			& WAT$^{**}$ (Piccinno and Ferragina, 2014) & 0.72 & 0.55 & 0.62 & 0.68 & 0.42 & 0.52\\
			& Spotlight$_{0.1}$ (Daiber et al., 2013)	& 0.73	& 0.76	& 0.75	& 0.69	& 0.73	& 0.71\\		
			& OLLIE$^{*}$ (Schmitz et al., 2012) & 0.45	& 0.19 &	0.27 & 0.44	& 0.18 &	0.26\\
			& AIDA (Yosef et al., 2011)	& {\bf 0.8}	& 0.6	& 0.68	& 0.77	& 0.45	& 0.57\\\hline
			(A)	& DSA$_{(60,0)}$	& 0.68	& 0.75	& 0.71	& 0.65	& 0.72	& 0.68\\
			(B)	& ${PG}_{(3L,80K)}(DSA_{DICT})$	& 0.71	& 0.74	& 0.73	& 0.68	& 0.72	& 0.7\\
			(C)	& ${PG}_{(2L,18K)}(DSA_{(60,0)})$ &	0.71 &	0.81 &	0.76 &	0.68 &	0.76 &	0.72\\\hline
			(D)	& (B) + (C)	& 0.76	& 0.84 &	0.8	& 0.72	& 0.8 &	0.76\\
			(H)	& (C) + Spotlight$_{0.1}$	& 0.78	& {\bf 0.85} &	{\bf 0.81}	& 0.75	& {\bf 0.81} &	\textbf{0.78}\\
			\hline
	\end{tabular}}
\end{table}
				
To compare the performance of our model with state-of-the-art NER, we applied it to two common public datasets for NER (CoNLL-2003 and GENIA). 
Table \ref{tab:eval_conll} shows the results on the CoNLL-2003 dataset for two variants of our model (Setups B and C) trained on our large training set, without any further NER adaptation, as well as for their updated versions (Setups I, J, and K), which were fine-tuned with the training set of the shared task $CoNLL_{T}$, contrasted with the results of the two genuine state-of-the-art NE recognizers \cite{lample2016neural} and \cite{devlin-etal-2019-bert} and DBpedia Spotlight. It should be noted that NER is a concept extraction subtask which aims at detecting less generic concepts. Consider the following statistics for the clarity: from about 69K nouns in the CoNLL-2003 training set, only 31K nouns are part of NEs (e.g., S\&P, BAYERISCHE VEREINSBANK, London Newsroom, Lloyds Shipping Intelligence Service), while the remaining 38K nouns (as in ``air force'', ``deposit rates'', ``blue collar workers'') are not part of NEs; as for GENIA, from about 132K nouns, only 93K form NEs (e.g., ``tumor necrosis factor'', ``terminal differentiation'', ``isolated polyclonal B lymphocytes''), while the remaining 39K do not (as in ``colonies'', ``interpretation'', ``notion'', ``circular dichroism'', ``differential accumulation'').
		
\begin{table}
	\centering
	\caption{\label{tab:eval_conll}Results on the CoNLL-2003 datasets}
	\small
	\resizebox{\columnwidth}{!}{\begin{tabular}{|l|l|c|c|c|c|c|c|}
			\hline
			\multicolumn{2}{|c|}{} & \multicolumn{3}{c|}{CoNLL-2003 (test-a)} & \multicolumn{3}{c|}{CoNLL-2003 (test-b)}\\\hline
			Setup & Model 	& $P$	& $R$	& $F_{1}$ & $P$ & $R$	& $F_{1}$\\\hline
			& BERT NER (Delvin et al., 2019) & 0.95 & 0.96 & 0.95 & 0.94 & 0.94 & 0.94\\
			& NER Tagger (Lample et al., 2016) & 0.97 & 0.97 & \textbf{0.97} & 0.97 & 0.96 & \textbf{0.96}\\
			& Spotlight$_{0.5}$ (Daiber et al., 2011) & 0.9 & 0.63 & 0.74 & 0.9 & 0.65 & 0.75\\
			& Spotlight$_{0.1}$ (Daiber et al., 2011) & 0.77 & 0.77 & 0.77 & 0.76 & 0.77 & 0.77\\\hline
			(B) & ${PG}_{(3L,80K)}(DSA_{DICT})$ & 0.81 & 0.78 & 0.8 & 0.81 & 0.79 & 0.8\\
			(C) & ${PG}_{(2L,18K)}(DSA_{(60,0)})$ & 0.82  & 0.82 & 0.82 & 0.79  & 0.81 & 0.8\\\hline
			(I) & $FineTune((B), CoNLL_{T})$ & 0.95 & 0.92 & 0.93 & 0.95 & 0.92 & 0.94\\
			(J) & $FineTune((C), CoNLL_{T})$ & 0.94 & 0.91 & 0.93 & 0.96  & 0.92 & 0.94\\
			(K) & (I) + (J) & 0.94 & 0.93 & 0.93 & 0.95 & 0.93 & 0.94\\
			\hline
	\end{tabular}}
\end{table}
			
\noindent Table \ref{tab:eval_genia} shows the results of our models fine-tuned with GENIA along with the results of concept identification by the recently published model \cite{strakova2019neural},\footnote{https://github.com/ufal/acl2019\_nested\_ner} which provides the most promising scores on different GENIA tasks.
			
\begin{table}
	\centering
	\caption{\label{tab:eval_genia}Results on the GENIA dataset}
	\small
	\resizebox{0.6\columnwidth}{!}{\begin{tabular}{|l|l|c|c|c|}
			\hline
			\multicolumn{2}{|c|}{} & \multicolumn{3}{c|}{GENIA} \\\hline
			Setup & Model 	& $P$	& $R$	& $F_{1}$ \\\hline
			& seq2seq (Strakov\'a et al., 2019) & 0.86 & 0.79 & 0.82\\\hline
			(L) & $FineTune((B), GENIA_{T})$ & 0.85 & 0.8 & 0.82\\
			(M) & $FineTune((C), GENIA_{T})$ & 0.84 & 0.77 & 0.81\\
			(N) & (L) + (M) & 0.85 & 0.8 & \textbf{0.83}\\
			\hline
	\end{tabular}}		
\end{table}
				
\subsection{Discussion}
\label{sec:discuss}
				
Tables \ref{tab:eval_small} and \ref{tab:eval_large} show that a combination of the different variants of the proposed pointer-generator model, which do not rely on external dictionaries after being trained (cf. Setup D), outperforms in terms of recall and $F_{1}$-score nearly all other models, including the dictionary lookup-based DBpedia Spotlight, which is a hard to beat as it was applied to ``known'' data. However, a combination of the pointer--generator model with DBpedia Spotlight is even better; it outperforms DBpedia Spotlight by 10\%. In other words, a deep model combined with a DBpedia-lookup is the best solution for generic CE. This applies to both runs displayed in Table \ref{tab:eval_large}, while it is to be noted that all tested models show a lower performance in the discovery of non-named entities. In particular, the NER models expectedly suffer a dramatic drop in recall. As far as precision is concerned, DBpedia Spotlight on its own is considerably better than any other proposal on the two small domain-specific test sets, while AIDA is best on the open domain test set. This is to be expected for dictionary lookup-based strategies. Also, as to be expected, DBpedia Spotlight, applied with its confidence coefficient equal to 0.1, showed significantly better recall than with the default value of 0.5, although $F_{1}$-score was lower. The experiment on the CoNLL-2003 dataset shows that the proposed model for generic CE performs well even without any special adjustment ($F_{1} = 0.8$ -- $0.82$). It can be further fine-tuned to the specific dataset resulting in scores comparable to state-of-the-art, even if not designed specifically for the NER task ($F_{1} = 0.93$ -- $0.94$), while its overall CE performance is better than of the targeted NER models (compare, e.g., (B)+(C) with Lample et al. (2016)'s NER in Tables \ref{tab:eval_small} and \ref{tab:eval_large}.

\section{Conclusions}
\label{sec:concls}
				
We presented an adaptation of the pointer--generator network model \cite{see2017get} to generic open-domain concept extraction. Due to its capacity to cope with OOV concept labels, it outperforms dictionary lookup-based CE such as DBpedia Spotlight or AIDA in terms of recall and $F_{1}$-score. It also shows an advantage over deep models that focus on NER only since it also covers non-named concept categories. However, a combination of the pointer--generator model with DBpedia Spotlight seems to be the best solution since it takes advantage of both the neural model and the dictionary lookup.
In order to facilitate a solid evaluation of the proposed model and compare it to a series of baselines, we utilized Wikipedia pages with text snippet links as a sparsely concept-annotated dataset. To ensure that our model is capable of extracting all generic concepts instead of detecting only texts of the page links, we ignored this sparse annotation during training. Instead, we compiled a large densely concept-annotated dataset for leveraging it within the distant supervision using the algorithm described above. To the best of our knowledge, no such dataset was available so far.
In the future, we plan to address the problem of multilingual concept extraction, using pre-trained multi-lingual embeddings and compiling another large dataset that contains a higher percentage of non-named entity concepts.

The code for running our pretrained models is available in the following GitHub repository: \url{https://github.com/TalnUPF/ConceptExtraction/}.
				
\section{Acknowledgments} The work presented in this paper has been supported by the European Commission within its H2020 Research Programme under the grant numbers 700024, 700475, 779962, 786731, 825079, and 870930.

%
%
%
%
  \bibliography{Shvets_Wanner_EKAW2020_submission_accepted}
  \bibliographystyle{splncs04}
\end{document}